\documentclass{article}%
\usepackage{graphicx}
\usepackage{amsmath}
\usepackage{amsfonts}
\usepackage{amssymb}
\usepackage{graphicx}%
\providecommand{\U}[1]{\protect\rule{.1in}{.1in}}
\begin{document}

\title{Gradient boosting machine with partially randomized decision trees}
\author{Andrei V. Konstantinov and Lev V. Utkin\\Peter the Great St.Petersburg Polytechnic University (SPbPU)\\St.Petersburg, Russia\\e-mail: andrue.konst@gmail.com, lev.utkin@gmail.com}
\date{}
\maketitle

\begin{abstract}
The gradient boosting machine is a powerful ensemble-based machine learning
method for solving regression problems. However, one of the difficulties of
its using is a possible discontinuity of the regression function, which arises
when regions of training data are not densely covered by training points. In
order to overcome this difficulty and to reduce the computational complexity
of the gradient boosting machine, we propose to apply the partially randomized
trees which can be regarded as a special case of the extremely randomized
trees applied to the gradient boosting. The gradient boosting machine with the
partially randomized trees is illustrated by means of many numerical examples
using synthetic and real data.


\end{abstract}

\section{Introduction}

Ensemble-based techniques can be regarded as one of the most efficient ways to
improve machine learning models. The basic idea behind ensemble-based
techniques is to combine the base or weak classifiers or regressors
\cite{Ferreira-Figueiredo-2012,Kuncheva-2004,Polikar-2012,Rokach-2019,Sagi-Rokach-2018,Wozniak-etal-2014,ZH-Zhou-2012}%
. The most well-known and efficient ensemble-based techniques are random
forests \cite{Breiman-2001} as a combination of the bagging
\cite{Breiman-1996} and the random subspace \cite{Ho-1998} methods, Adaboost
\cite{Freund-Shapire-97}, the gradient boosting machines (GBMs)
\cite{Friedman-2001,Friedman-2002} and its modifications XGBoost
\cite{Chen-Guestrin-2016}, LightGBM \cite{Guolin-etal-17}, CatBoost
\cite{Dorogush-etal-2018}.

GBMs have illustrated their efficiency for solving regression problems.
According to the technique, the first iteration starts from the guessed
prediction, then residuals are calculated as differences between guessed
predictions and target variables. The residuals instead of target variables
form a new dataset such that the next base model is built on the dataset. A
regression tree is often used to predict new residuals. The GBM iteratively
computes the sum of all previous regression tree predictions and updating
residuals to reflect changes in the model. As a result, a set of regression
trees is built in the GBM such that each successive tree predicts the
residuals of the preceding trees given an arbitrary differentiable loss
function \cite{Sagi-Rokach-2018}. An interesting modification of the GBM on
the basis of the so-called deep forests \cite{Zhou-Feng-2017a} is the
multi-layered gradient boosting decision tree model \cite{Feng-Yu-Zhou-2018}.
Another modification is the soft GBM \cite{Feng-etal-20}.

In order to improve ensemble-based models using decision trees, Geurts et al.
\cite{Geurts-etal-06} proposed to apply the individual extremely randomized
trees which aim to generate a decision forest while injecting randomness.
According to the extremely randomized trees, the best splitting feature is
selected from a random subset of features. Cut-points splitting features are
also randomly selected during training the trees. The possible high bias and
variance of the trees can be canceled by fusing a sufficiently large forest of
trees \cite{Geurts-etal-06}.

Various types of randomized trees have been developed and used in
ensemble-based models. A similar idea has been implemented for a randomized
C4.5 decision tree \cite{Dietterich-00}. Instead of selecting the best
attribute at each stage, it selects an attribute from the set of the best $m$
features randomly with equal probability. Cutler and Zhao
\cite{Cutler-Zhao-01} proposed the PERT (Perfect Ensemble Random Trees)
approach for building perfect-fit classification trees with random split
selection. Every base model in the PERT ensemble randomly chooses both the
feature on which to split and the split itself.

It should be noted that one of the drawbacks of the GBM is its computational
complexity. Therefore, there are many attempts to reduce it by introducing
various randomization schemes. For example, Lu and Mazumder
\cite{Lu-Mazumder-18} proposed the Randomized GBM which leads to significant
computational gains compared to the GBM by using a randomization scheme to
reduce the search in the space of weak learners. Lu et al. \cite{Lu-etal-2019}
proposed the Accelerated GBM by incorporating Nesterov's acceleration
techniques into the design of the GBM.

Another difficulty of using the GBM is a possible discontinuity of the
regression function, which arises when some regions of training data are not
densely covered by points, for example, when the training set is very small.
We illustrate by examples, that the standard GBM as well as the extremely
randomized trees may lead to incorrect prediction and the discontinuity of the
regression function.

In order to overcome the problems of the computational complexity as well as
the discontinuity of the regression function, we propose the GBM with
partially randomized trees as base learners. In contrast to extremely
randomized trees, the cut-point for partitioning each feature in these trees
is determined randomly from a uniform distribution, but the best feature is
selected such that it maximizes the score. On the one hand, the partially
randomized trees can be regarded as a special case of extremely randomized
trees. On the other hand, their incorporation into the GBM differ them from
the extremely randomized trees. Many numerical examples show that the GBM with
partially randomized trees provide better prediction results in comparison
with many ensemble-based models, including extremely randomized trees.
Moreover, the corresponding GBMs are computationally simpler in comparison
with the original GBM.

\section{Regression problem statement}

The standard regression problem can be stated as follows. Given $N$ training
data (examples, instances, patterns) $D=\{(x_{1},y_{1}),...,(x_{N},y_{N})\}$,
in which $x_{i}$ belongs to a set $\mathcal{X}\subset\mathbb{R}^{m}$ and
represents a feature vector involving $m$ features, and $y_{i}\in\mathbb{R}$
represents the observed output or the target value such that $y_{i}%
=f(x_{i})+\varepsilon$. Here $\varepsilon$ is the random noise with
expectation $0$ and unknown finite variance. Machine learning aims to
construct a regression model or an approximation $g$ of the function $f$ that
minimizes the expected risk or the expected loss function
\begin{equation}
L(f)=\mathbb{E}_{(x,y)\sim P}~l(y,g(x))=\int_{\mathcal{X}\times\mathbb{R}%
}l(y,g(x))\mathrm{d}P(x,y), \label{Imp_SVM16}%
\end{equation}
with respect to the function parameters. Here $P(x,y)$ is a joint probability
distribution of $x$ and $y$; the loss function $l(\cdot,\cdot)$ may be
represented, for example, as follows:
\begin{equation}
l(y,g(x))=\left(  y-g(x)\right)  ^{2}.
\end{equation}

There are many powerful machine learning methods for solving the regression
problem, including regression random forests
\cite{Biau-Scornet-2016,Breiman-2001}, the support vector regression
\cite{Smola-Scholkopf-2004}, etc. One of the powerful methods is the GBM
\cite{Friedman-2002}, which will be considered below.

\section{A brief introduction to the GBM}

Let us consider the gradient boosting decision tree algorithm
\cite{Friedman-2002}. The algorithm is an iterative construction of a model as
an ensemble of base (weak) prediction models built in a stage-wise fashion
where each base model is constructed, based on data obtained using an ensemble
of models already built on previous iterations, as an approximation of the
loss function derivative. A model of size $M$ is a linear combination of $M$
base models:%

\begin{equation}
g_{M}(x)=\sum_{i=0}^{M}\gamma_{i}h_{i}(x), \label{grad_bost_10}%
\end{equation}
where $h_{i}$ is the $i$-th base model; $\gamma_{i}$ is the $i$-th coefficient
or the $i$-th base model weight.

The gradient boosting algorithm can be represented as the following steps:

\begin{enumerate}
\item Initialize the zero base model $h_{0}(x)$, for example, with the
constant value.

\item Calculate the residual $r_{i}^{(t)}$ as a partial derivative of the
expected loss function $L(x_{i},y_{i})$ at every points of the training set,
$i=1,...,N$.

\item Build the base model $h_{t}(x)$ as regression on residuals
$\{(x_{i},r_{i}^{(t)})\}$;

\item Find the optimal coefficient $\gamma_{t}$ at $h_{t}(x)$ regarding the
initial expected loss function (\ref{grad_bost_20});

\item Update the whole model $g_{t}(x)=g_{t-1}(x)+\gamma_{t}h_{t}(x)$;

\item If the stop condition is not fulfilled, go to step 2.
\end{enumerate}

Here the loss function depends on the machine learning problem solved
(classification or regression). Suppose that $(M-1)$ steps produce the model
$g_{M-1}(x)$. For constructing the model $g_{M}(x)$, the model $h_{M}(x)$ has
to be constructed, i.e., there holds
\begin{equation}
g_{M}(x)=\sum_{t=1}^{M}\gamma_{t}h_{t}(x)=g_{M-1}(x)+\gamma_{M}h_{M}(x).
\end{equation}

The dataset for constructing the model $h_{M}(x)$ is chosen in such a way as
to approximate the expected loss function partial derivatives with respect to
the function of the already constructed model $g_{M-1}(x)$. Let us denote
residuals $r_{i}^{(M)}$ defined as the values of the loss function partial
derivative at point $g_{M-1}(x_{i})$ in the current iteration $M$,%
\begin{equation}
r_{i}^{(M)}=-\left.  \frac{\partial L(z,y_{i})}{\partial z}\right\vert
_{z=g_{M-1}(x_{i})}.
\end{equation}

By using the residuals, a new training set $D_{M}$ is derived as follows:%
\begin{equation}
D_{M}=\left\{  \left(  x_{i},r_{i}^{(M)}\right)  \right\}  _{i=1}^{N},
\end{equation}
and the model $h_{M}$ can be constructed on $D_{M}$ by solving the following
optimization problem
\begin{equation}
\min\sum_{i=1}^{N}\left\Vert h_{M}(x_{i})-r_{i}^{(M)}\right\Vert ^{2}.
\label{grad_bost_19}%
\end{equation}

Hence, an optimal coefficient $\gamma_{M}$ of the gradient descent can be
obtained as:%
\begin{equation}
\gamma_{M}=\arg\min_{\gamma}\sum_{i=1}^{N}L\left[  g_{M-1}(x)+\gamma
h_{M}(x_{i}),y_{i}\right]  . \label{grad_bost_20}%
\end{equation}

Then we get the following model at every point $x_{i}$ of the training set
\begin{equation}
g_{M}(x)=g_{M-1}(x)+\gamma_{M}h_{M}(x)\approx g_{M-1}(x)-\gamma_{M}\left.
\frac{\partial L(z,y_{i})}{\partial z}\right\vert _{z=g_{M-1}(x_{i})}.
\end{equation}

The above algorithm minimizes the expected loss function by using decision
trees as base models. Its parameters include depths of trees, the learning
rate, the number of iterations. They are selected to provide a high
generalization and accuracy depending on an specific task.

The gradient boosting algorithm is a powerful and efficient tool for solving
regression problems, which can cope with complex non-linear function
dependencies \cite{Natekin-Knoll-13}.

\section{Improved gradient boosting algorithm}

\subsection{Motivation}

Let us consider a peculiarity of GBMs constructed using decision trees. It is
easy to show that a function having the GBM structure allows us to approximate
any continuous function using decision trees with the number of decision rules
equal to the number of features. Obviously, if we would not restrict the
number of decision rules, then we can approximate any continuous function by
constructing only a single tree. On the other hand, if all features are
significant, then, in a general case, it is impossible to approximate the
function by a linear combination of decision trees with the number of decision
rules less than the number of features. Indeed, if it were possible, then any
function of several variables could be decomposed into the sum of functions of
one variable, but this is incorrect.

A regression trees aims to recursively split the training set with binary
tests in the form $x_{i}\leq t$, where $x_{i}$ is one of the input variables
or features and $t$ a threshold. A criterion for determining the splitting
parameters in the tree growing procedure is to reduce as much as possible the
variance of $y$ in the two subsamples resulting from that split. The
regression tree can be seen as a kind of additive model of the form
\cite{Hastie-Tibshirani-1990,Hastie-Tibshirani-Friedman09}:
\begin{equation}
tree(x)=\sum_{l\in V}b_{l}\cdot\mathbb{I}[x\in R_{l}], \label{grad_bost_9}%
\end{equation}
where $\mathbb{I}[\cdot]$ is the indicator function taking the value $1$ if
its argument is true and $0$ otherwise; $b_{l}$ is the value in the $l$-th
leaf; $R_{l}$ is the region defined by disjoint partitions of the training set
at the $l$-th leaf; $V$ is the set of leaves of the decision tree.

The approximation function $g$ from (\ref{grad_bost_10}) can be rewritten by
using (\ref{grad_bost_9}) as follows:%
\[
g_{M}(x)=\sum_{i=0}^{M}\gamma_{i}\sum_{l\in V_{i}}b_{l}\cdot\mathbb{I}[x\in
R_{l}],
\]
where $V_{i}$ is the set of leaves of the $i$-th tree.

The indicator function $\mathbb{I}[x\in R_{l}]$ is determined through all
decision rules corresponding to the $l$-th leaf:
\[
\mathbb{I}[x\in R_{l}]=\left(
{\displaystyle\prod\limits_{j\in Left(l)}}
\mathbb{I}\left[  x_{j}\leq t_{j}^{(left)}\right]  \right)  \cdot\left(
{\displaystyle\prod\limits_{j\in Right(l)}}
\mathbb{I}\left[  x_{j}>t_{j}^{(right)}\right]  \right)  ,
\]
where $Left(l)$ is the set of true rules leading to the $l$-th leaf;
$Right(l)$ is the set of false rules leading to the $l$-th leaf; $t_{j}$ is
the threshold of the $j$-th rule.

If the number of rules corresponding to each feature in every tree is 1, then
such trees correspond to the angles of parallelepipeds whose faces are
hyperplanes parallel to all axes except for one axis. The entire domain of the
function is covered by means of such parallelepiped angles. In addition, the
entire domain of the function can be divided into equal cells with an
orthogonal grid using the sum of such trees, and a unique value can be
assigned to each cell. So, having selected the parameter values {}{}in a
certain way, we can approximate an arbitrary continuous function with a given accuracy.%

\begin{figure}
[ptb]
\begin{center}
\includegraphics[
height=2.1127in,
width=2.7916in
]%
{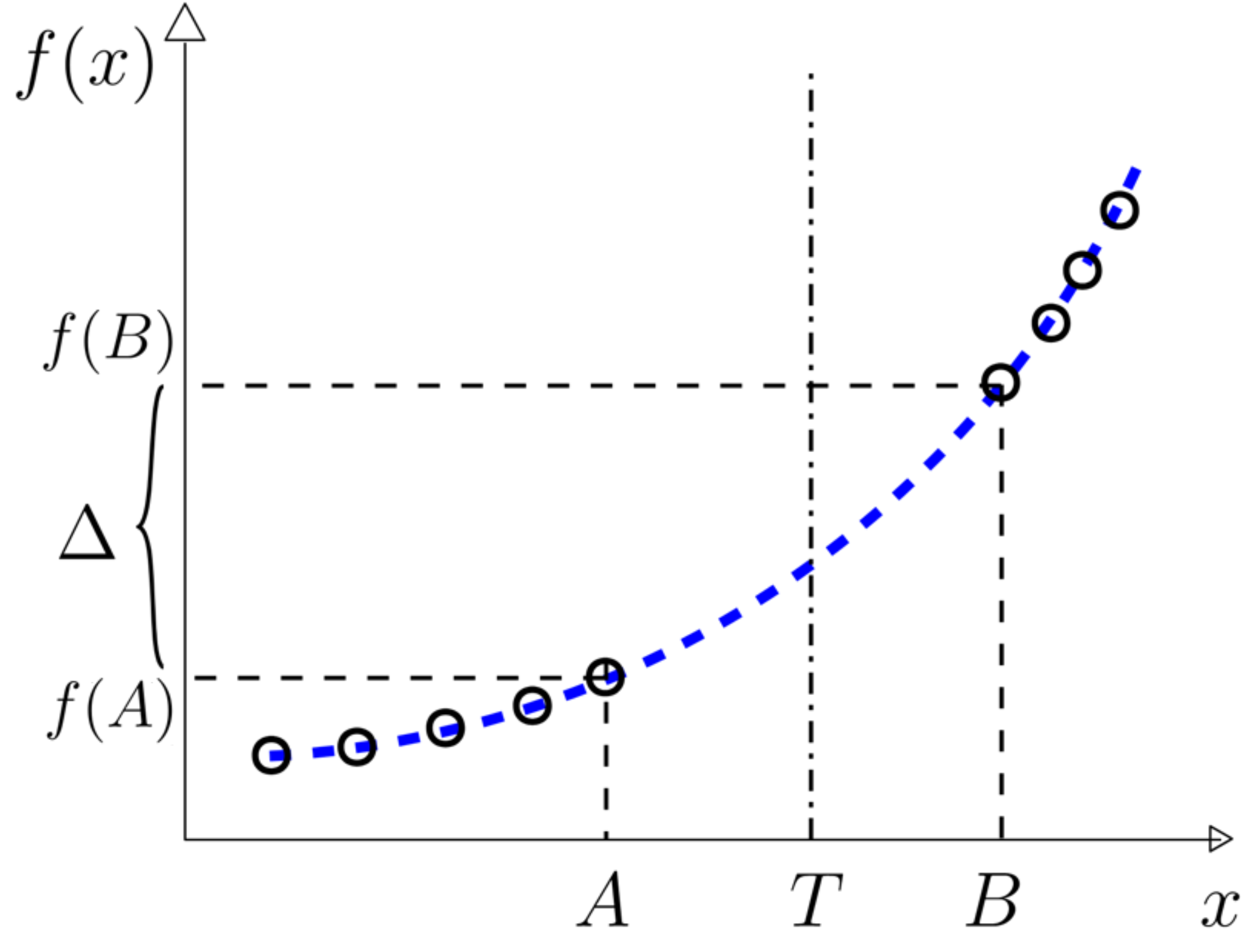}%
\caption{Different discontinuities of the function $f$}%
\label{f:func_gap}%
\end{center}
\end{figure}

In practice, observations of target values {}{}may be noisy. Moreover, some
parts of the function domain are often less densely covered with points of the
training set than others. When constructing a GBM on the basis of decision
trees, the GBM may have large discontinuities in regions of the absence of
points as it is shown in Fig. \ref{f:func_gap}. It can be seen from Fig.
\ref{f:func_gap} that the distance between the target values $f(A)$ and $f(B)$
{}{}of points $A$ and $B$ denoted as $\Delta$ is much larger than distances
between the target values {}{}of the remaining neighboring points to the left
of $A$ or to the right of $B$. After constructing the decision tree, distances
between predictions of points to the left of $T$ and to the right of $T$ will
be no less than $\Delta$. This is due to the algorithm for constructing the
decision tree: since the average values {}{}to the left and right of $T$ are
far from each other, then the locally optimal partition lies between $A$ and
$B$. As a rule, according to \cite{Breiman-etal-1984}, $T$ is chosen exactly
in the middle between $A$ and $B$. In the best case, if decision trees near
$A$ and $B$ will have the values {}{}of the points closest to $T$, that is,
the values {}{}at points $A$ and $B$, then the interval of the function
discontinuity predicted by the decision tree will be strictly equal to
$\Delta$, or it will be larger than $\Delta$. As a result, using the standard
algorithms for constructing decision trees, it is difficult to construct a
function that behaves in the discontinuity intervals in a more smooth way.
Moreover, if the region of training data is not densely covered by points,
then the loss function and the absolute value of its derivative near points
$A$ and $B$ will be close to zero even if the interval of discontinuity is
large. This is because there are no observations in the region between $A$ and
$B$. This implies that the predicted value for an example close to $T$ at the
next iterations of the GBM will be close to zero, and, therefore, the function
will not be smoothed. If the absolute value of the derivative is large enough,
for example, if the predictions of the previous models (iterations) in the GBM
are constant in this region, then this situation will be repeated, and the
function will contain a similar discontinuity interval even of a larger size.
Since there are no new points between $A$ and $B$, it is impossible to obtain
a division different from $T$ in this interval.

A continuous (or piecewise continuous) function can be approximated by a
piecewise constant function, having the structure of the GBM with decision
trees, on a uniform grid. However, many algorithms for building decision trees
like CART can build for each feature only the grid nodes located exactly in
the middle between pairs of neighboring points of the training set. It should
be noted that there are histogram-based methods \cite{Guryanov-19} that do not
necessarily build partitions exactly in the middle, but these methods consider
smaller numbers of nodes which are determined by points of the training set.
Therefore, the grid turns out to be larger and coarser in regions not densely
covered by the points of the training set. Since decision trees correspond to
piecewise constant functions, then large jumps are produced due to the rapid
growth of the approximated function in such regions. The jumps lead to an
increase of the loss. Moreover, the GBM does not reduce the loss because:

\begin{enumerate}
\item each new tree is not able to build partitions anywhere except in the
middle between pairs of neighboring points;

\item empirical loss does not take into account the error in the interval
between adjacent points.
\end{enumerate}

\subsection{Partially randomized decision trees and GBM}

\begin{figure}
[ptb]
\begin{center}
\includegraphics[
height=2.1837in,
width=2.8625in
]%
{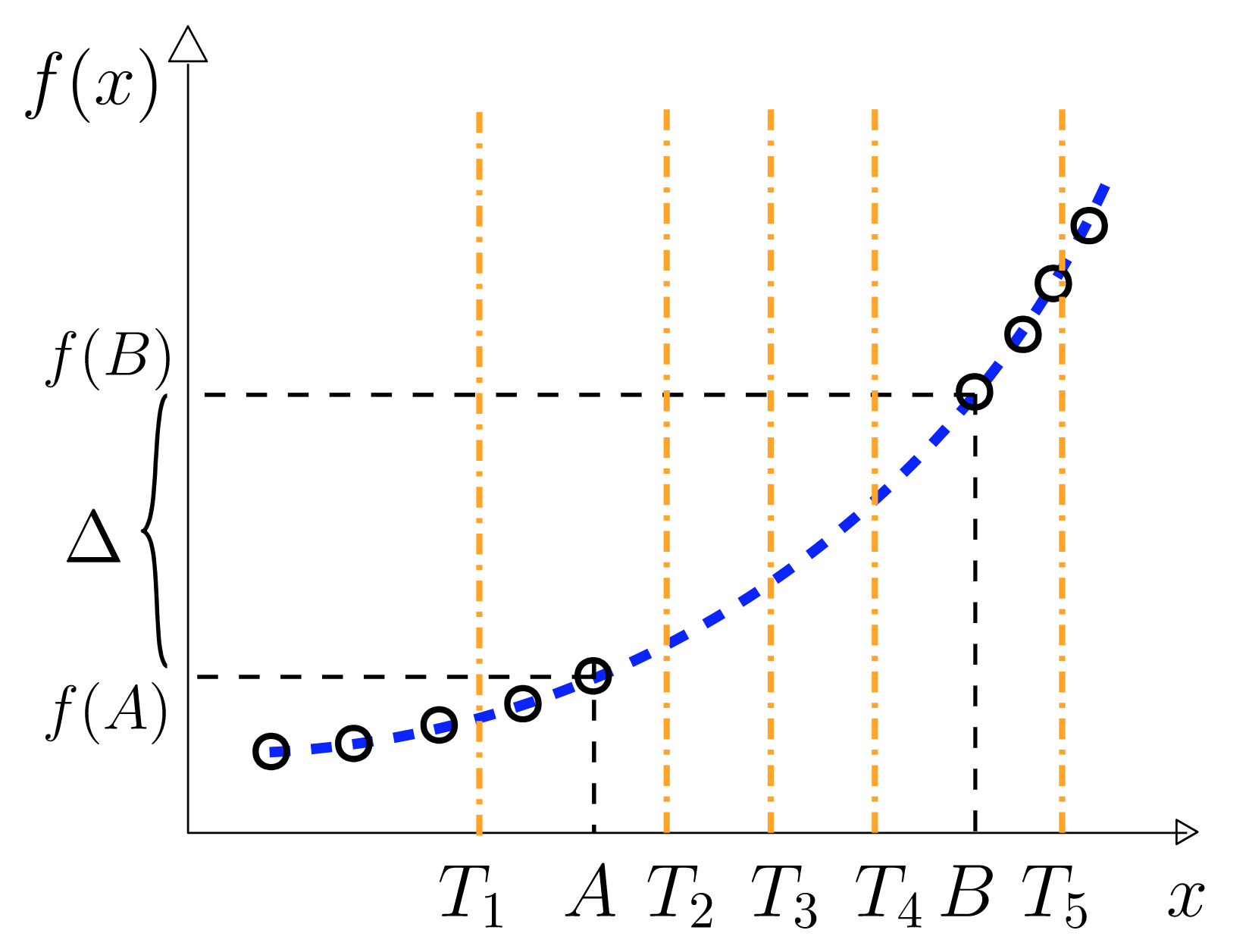}%
\caption{The Discontinuity of function $f $ and the regions obtained by
splitting}%
\label{f:func_gap_2}%
\end{center}
\end{figure}

In order to solve the aforementioned problem, it is enough to ensure the
construction of a denser grid, that is, to change the algorithm for building
decision trees so that thresholds of the partition rules are selected not only
from the set of midpoints of pairs of neighboring points of the training
dataset. A simple way is to replace the threshold of each partition with a
random one located within the interval corresponding to the threshold after
building a tree by means of a deterministic algorithm like CART. For example,
Fig. \ref{f:func_gap_2} shows how the partition $T$ is replaced by $T_{2}$,
$T_{3}$ or $T_{4}$, or any other values in the interval $(A,B)$. As a result,
partitions in the ensemble trees will be independent, and will allow covering
the interval more tightly. However, by taking into account the possible
overfitting of the GBM as well as its inability to parallelize the training
process which takes considerable time, we propose an approach which is not
based on updating partitions constructed by the deterministic algorithm and
has several advantages.

The basic idea behind the proposed approach is to use an algorithm for
constructing partially randomized decision trees as a special case of the
so-called extremely randomized trees \cite{Geurts-etal-06}. The extremely
randomized decision tree is grown by selecting at each node $K$\ random splits
such that a split includes the random choice of a feature $x_{i}$\ and the
random choice of a threshold $t_{i}$. One of the splits is fixed which
maximizes the score. If the parameter $K$\ is 1, then the corresponding trees
are extremely randomized ones. As indicated by Wehenkel et al.
\cite{Wehenkel-etal-06}, the tree model may be improved with respect to a
certain dataset by using larger values of $K$. Since we aim to compact the
grid, it makes sense to consider all the features at every stage of building a
tree, not limited to $K$ random features. Otherwise, a larger number of
iterations may be required to construct a sufficient number of partitioning
rules for each feature, and also to ensure the convergence if the approximated
function depends on the number of features exceeding $K$. According to the
partially randomized decision trees, the cut-point (threshold) for
partitioning each feature in these trees is determined randomly from a uniform
distribution. Then the best feature is selected such that it maximizes the
score, and the partition is performed by using this best feature in accordance
with the same random cut-point.

We call decision trees constructed by the well-known deterministic algorithms,
such as CART or C4.5, as deterministic trees below in order to distinguish
them from partially randomized trees. A single partially randomized tree is
worse on average (in the sense of the standard error) than a deterministic
decision tree with locally optimal cut-points of the same depth. Moreover, it
often has large jumps at places where the deterministic tree has smaller
jumps, for example, smaller than the average jump between two adjacent points.
However, ensembles of partially randomized trees in the form of the GBM have
better properties than ensembles of deterministic trees.

First, we consider splitting procedures for one feature for simplicity and
return to Fig. \ref{f:func_gap}. Every tree in the ensemble independently
determines its cut-point for a particular feature, perhaps more than once,
that is, it splits the domain of the function into several regions (intervals)
such that every region corresponds to a certain cut-point. Some of these
regions contain points to the left of $A$, some of the them contain points to
the right of $B$ (see Fig. \ref{f:func_gap_2}). The remaining part contains
points between $A$ and $B$ as it is depicted in Fig. \ref{f:func_gap_2}
(regions defined by intervals $[T_{2},T_{3}]$, $[T_{3},T_{4}]$). Regions of
the last remaining part allow filling the interval between $A$ and $B$ and
solve the problem of the function discontinuity.%

Thus, a single partially randomized decision tree does not allow solving the
aforementioned problem. However, the gradient boosting machine with partially
randomized decision trees allows us to construct smoother approximations of
functions. We will call the proposed model as a partially randomized gradient
boosting machine (PRGBM).

Let us point out two important advantages of the proposed PRGBM.

\begin{enumerate}
\item The PRGBM based on partially randomized decision trees has an enhanced
training adaptability in comparison with extremely randomized trees because
every subsequent partially randomized tree in the PRGBM is built with the aim
to improve the prediction of an already constructed ensemble, whereas
extremely randomized trees are built to approximate the same function.

\item An additional advantage of the PRGBM in comparison with the GBM based on
deterministic trees is a significant reduction of the learning time. This
follows from the fact that one partially randomized tree is built
approximately $N$ times faster than a deterministic tree because a single
partition is evaluated to construct a rule for each feature in the partially
randomized tree instead of $N$ partitions in the corresponding deterministic
tree. Hence, the construction time of the PRGBM in reduced in $N$ times.
\end{enumerate}

It is worth noting that the proposed approach has the following drawback: if
the objective function contains intervals of discontinuity, and points from
the training set cover regions around the intervals of discontinuity not
densely enough, then the GBM fills the intervals with intermediate values.

\begin{figure}
[ptb]
\begin{center}
\includegraphics[
height=1.5947in,
width=4.5299in
]%
{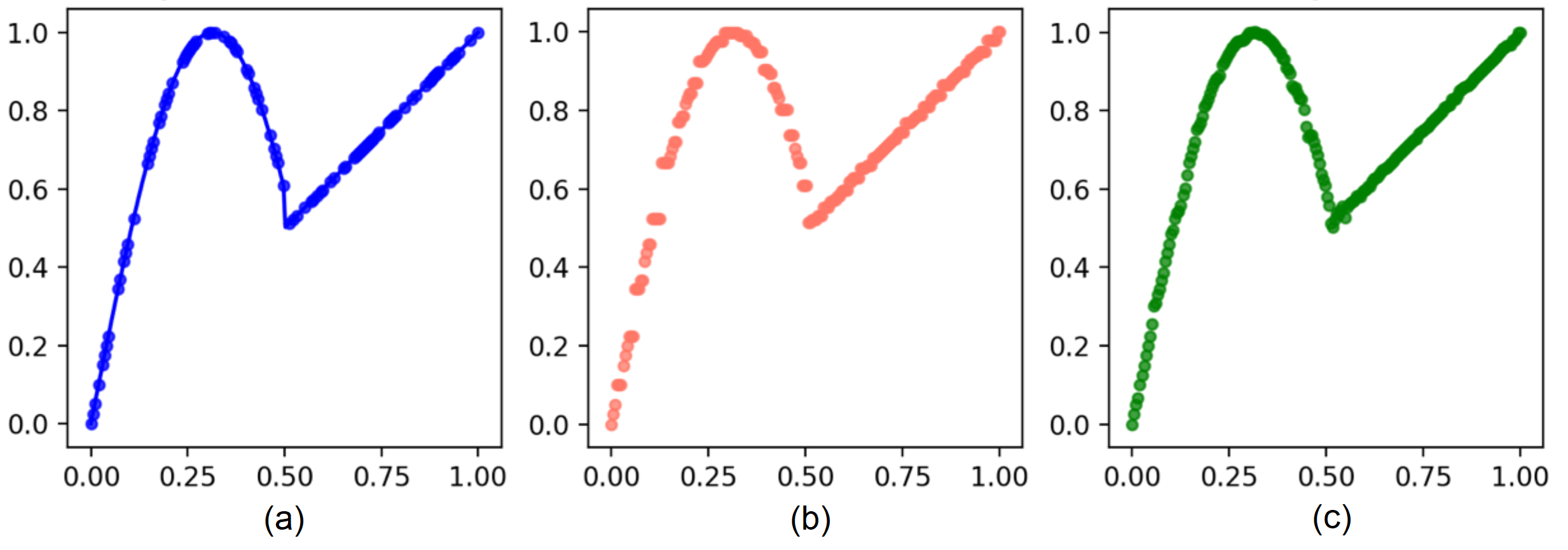}%
\caption{Training points (a), predictions of the GBM with deterministic trees
(b), and partially randomized trees (c)}%
\label{f:one_dim_func}%
\end{center}
\end{figure}

\begin{figure}
[ptb]
\begin{center}
\includegraphics[
height=1.9553in,
width=3.8372in
]%
{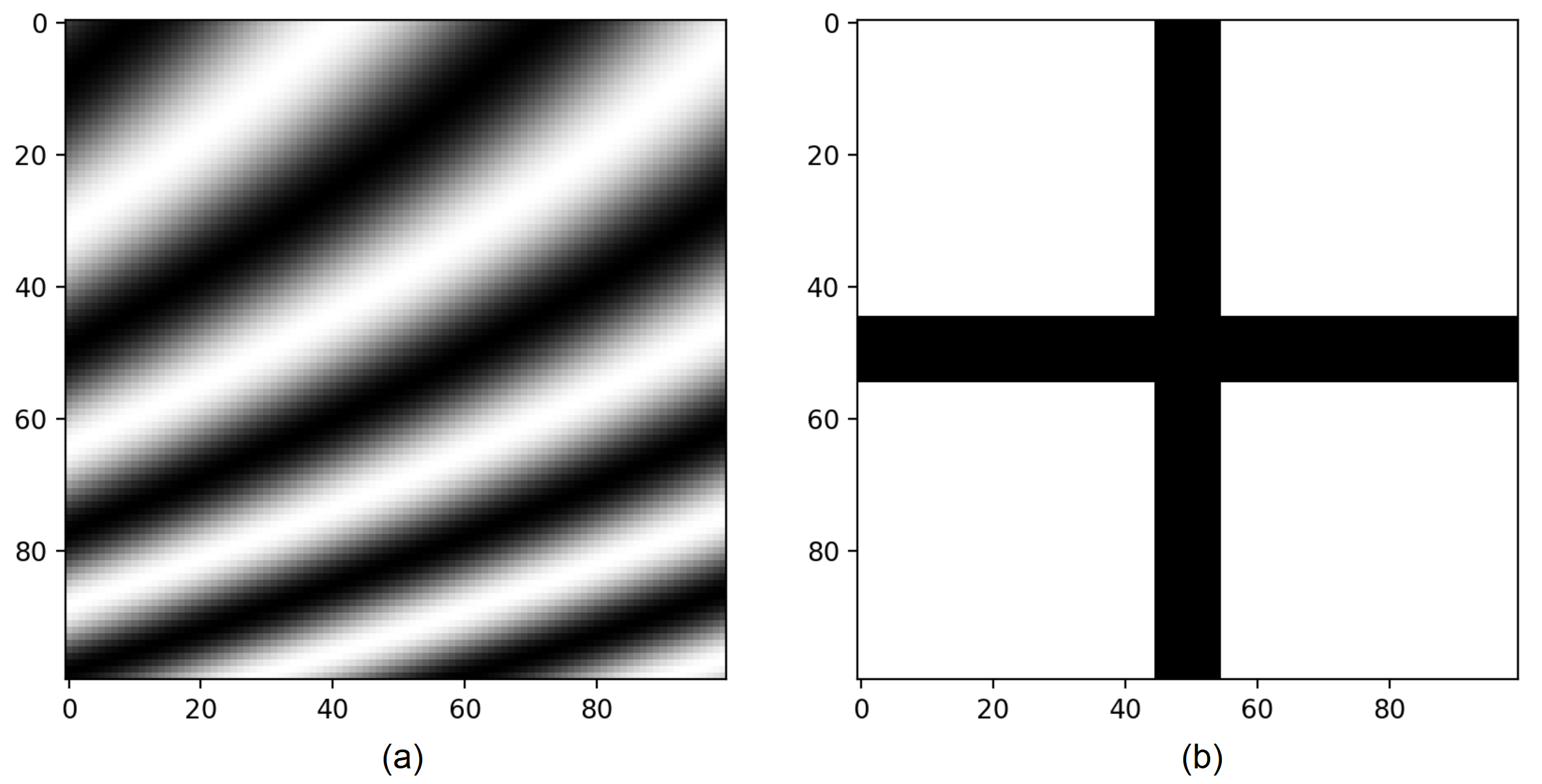}%
\caption{The original image (a) produced by the function $f$, and the cross
which is cut out of the image (b)}%
\label{f:two_dim_func_1}%
\end{center}
\end{figure}

\begin{figure}
[ptb]
\begin{center}
\includegraphics[
height=1.9182in,
width=5.5019in
]%
{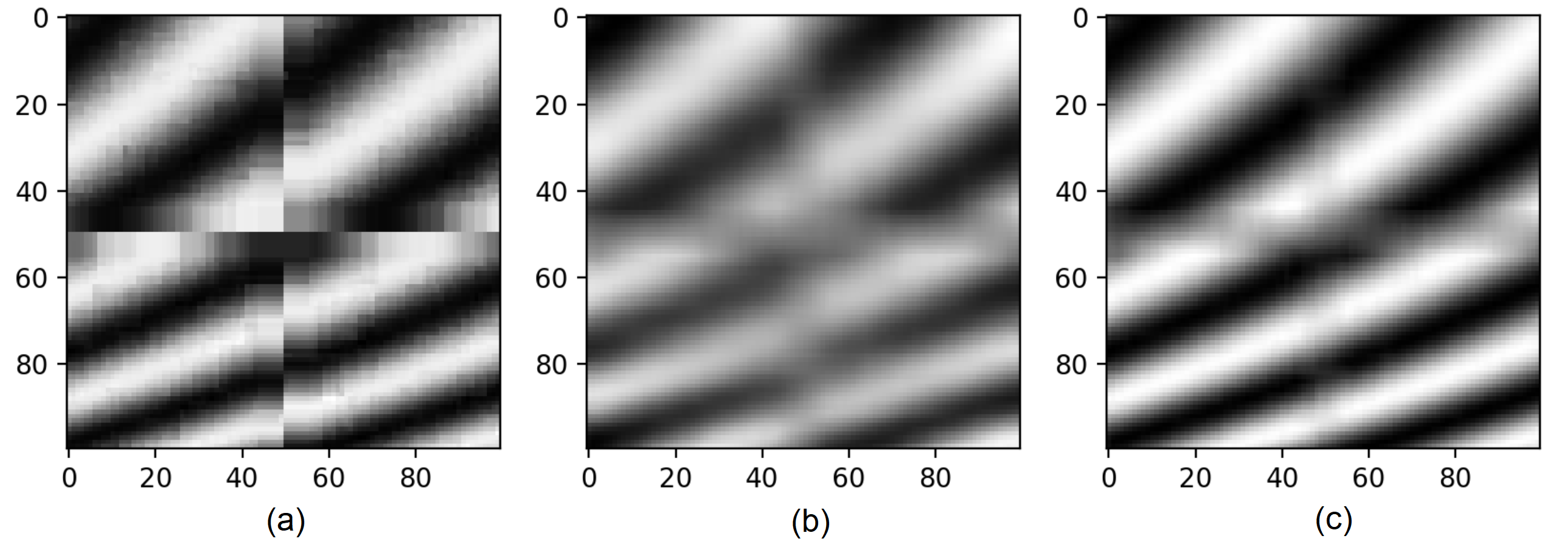}%
\caption{Predictions of the GBM with deterministic trees (a), extremely
randomized trees (b), and partially randomized trees (c)}%
\label{f:two_dim_func_2}%
\end{center}
\end{figure}

Let us consider two explanation examples with one-dimensional (Fig.
\ref{f:one_dim_func}) and two-dimensional (Figs. \ref{f:two_dim_func_1}%
-\ref{f:two_dim_func_2}) regression functions. The original (true) function
depicted by the solid line and points corresponding to examples from a
training set are shown in Fig. \ref{f:one_dim_func} (a). Predictions obtained
by means of the GBM based on deterministic trees and partially randomized
trees are depicted in the form of points in Fig. \ref{f:one_dim_func} (b) and
Fig. \ref{f:one_dim_func} (c), respectively. The depth of the trees in both
cases is the same and equal to 5. The horizontal axis in every figure
corresponds to feature values, the vertical axis shows target values.
Predictions are made on a uniform grid from 0 to 1 with 200 points. The
original function is defined as follows:
\[
f(x)=\sin(5x)\cdot\mathbb{I}\left[  x\leq\frac{1}{2}\right]  +x\cdot
\mathbb{I}\left[  x>\frac{1}{2}\right]  .
\]

It can be seen from Fig. \ref{f:one_dim_func} (a) that the insufficient
coverage by training examples of several regions of the feature negatively
affects the predictions of the GBM with deterministic trees whereas
predictions of the GBM with partially randomized trees successfully fill these
regions. One can clearly see in Fig.\ref{f:one_dim_func} (c) how predictions
the GBM with partially randomized trees fill the discontinuity interval in the
neighborhood of $0.5$.%

The two-dimensional function $f(x,y)$ is shown in Fig. \ref{f:two_dim_func_1}.
The function $f$ is defined on the interval from 0 to 1 as follows:%
\[
f(x,y)=\sin(10\cdot(x+\exp(1.1\cdot y))).
\]
It is depicted for convenience in the form of an image in Fig.
\ref{f:two_dim_func_1} (a), where the horizontal and vertical axes correspond
to the first $x$ and the second $y$ features, respectively, and the brightness
corresponds to the function values $f(x,y)$. The original image consists of
$100^{2}$ points. The training set is composed as follows. A cross shown in
Fig. \ref{f:two_dim_func_1} (b) is cut out of the image. Then a half of all
remaining (not cut) points in the image is taken as the training set. In other
words, the training set is a part of points in Fig. \ref{f:two_dim_func_1} (a)
without points belonging to the cross in Fig. \ref{f:two_dim_func_1} (b). The
discontinuity of the function $f$ arises from the cross.%

Predictions of three different models with trees of the same depth $9$ are
presented in Fig. \ref{f:two_dim_func_2}. Fig. \ref{f:two_dim_func_2} (a)
depicts results of the GBM with deterministic trees. Predictions of the GBM
with extremely randomized trees are shown in Fig. \ref{f:two_dim_func_2} (b).
Predictions of the GBM with partially randomized trees are depicted in Fig.
\ref{f:two_dim_func_2} (c). The number of basic models in each ensemble is
equal to $1000$ for all cases. It can be seen from Fig. \ref{f:two_dim_func_2}
(a) that the GBM with deterministic trees fills points of the cutout cross
with values {}{}of the nearest points which are available in the training set.
It is interesting to note that the GBM with extremely randomized trees Fig.
\ref{f:two_dim_func_2} (b) as well as the GBM with partially randomized trees
Fig. \ref{f:two_dim_func_2} (c) fill the cross with more appropriate values
{}{}than the GBM with deterministic trees. At the same time, the quality of
the GBM with partially randomized trees is noticeably better than that of the
GBM with extremely randomized trees.%

\section{Numerical experiments}

In order to study the proposed approach for solving regression problems, we
apply datasets described in Table \ref{t:regres_datasets} where the
abbreviation of every datasets, the corresponding number of examples $n$, and
the number of features $m$ are shown. The datasets are taken from open
sources, in particular, datasets California, Boston, and Diabetes can be found
in the corresponding R Package \textquotedblleft StatLib\textquotedblright;
the dataset HouseART can be found in the Kaggle platform; synthetic datasets
Friedman 1, 2, 3 are described at site:
https://www.stat.berkeley.edu/\symbol{126}breiman/bagging.pdf; datasets
Regression and Sparse are available in the Python Package \textquotedblleft
Scikit-Learn\textquotedblright. Table \ref{t:regres_datasets} is a brief
introduction about these datasets, while more detailed information can be
found from, respectively, the data resources.%

\begin{table}[tbp] \centering
\caption{A brief introduction about the regression data sets}%
\begin{tabular}
[c]{cccc}\hline
Data set & Abbreviation & $m$ & $n$\\\hline
California housing dataset & California & $8$ & $20640$\\\hline
House Prices: Advanced Regression Techniques & HouseART & $79$ &
$1460$\\\hline
ML housing dataset & Boston & $13$ & $506$\\\hline
Diabetes & Diabetes & $10$ & $442$\\\hline
Friedman 1 & Friedman 1 & $10$ & $100$\\\hline
Friedman 2 & Friedman 2 & $4$ & $100$\\\hline
Friedman 3 & Friedman 3 & $4$ & $100$\\\hline
Scikit-Learn Regression & Regression & $100$ & $100$\\\hline
Scikit-Learn Sparse uncorrelated & Sparse & $10$ & $100$\\\hline
\end{tabular}
\label{t:regres_datasets}%
\end{table}%

To evaluate the average accuracy, we perform a cross-validation with $100$
repetitions, where in each run, we randomly select $n_{\text{tr}}=3n/4$
training data and $n_{\text{test}}=n/4$ testing data. Different values for the
tuning parameters have been tested, choosing those leading to the best results.

Numerical results in the form of the mean squared errors for the regression
datasets are shown in Table \ref{t:regression_results}. The best performance
for each dataset is shown in bold. We compare the following six models: random
forest (RF), extremely randomized trees (ERT), GBM, CatBoost
\cite{Dorogush-etal-2018}, XGBoost \cite{Chen-Guestrin-2016}, PRGBM. It can be
seen from Table \ref{t:regression_results} that the proposed model provides
better results for $8$ datasets from $9$ ones.%

\begin{table}[tbp] \centering
\caption{Comparison of six models on the  regression datasets}%
\begin{tabular}
[c]{ccccccc}\hline
Data set & RF & ERT & GBM & CatBoost & XGBoost & PRGBM\\\hline
California & $2.56\times10^{-1}$ & $2.49\times10^{-1}$ & $2.08\times10^{-1}$ &
$2.20\times10^{-1}$ & $2.06\times10^{-1}$ & $\mathbf{2.04\times10}^{-1}%
$\\\hline
HouseART & $9.80\times10^{8}$ & $9.65\times10^{8}$ & $9.53\times10^{8}$ &
$9.15\times10^{8}$ & $9.84\times10^{8}$ & $\mathbf{8.98\times10}^{8}$\\\hline
Boston & $1.22\times10^{1}$ & $1.14\times10^{1}$ & $\mathbf{1.05\times10}^{1}$
& $1.07\times10^{1}$ & $1.12\times10^{1}$ & $1.11\times10^{1}$\\\hline
Diabetes & $3.37\times10^{3}$ & $3.24\times10^{3}$ & $3.29\times10^{3}$ &
$3.55\times10^{3}$ & $3.28\times10^{3}$ & $\mathbf{3.11\times10}^{3}$\\\hline
Friedman 1 & $1.06\times10^{1}$ & $8.87\times10^{0}$ & $7.23\times10^{0}$ &
$8.49\times10^{0}$ & $7.07\times10^{0}$ & $\mathbf{4.09\times10}^{0}$\\\hline
Friedman 2 & $5.84\times10^{3}$ & $1.48\times10^{3}$ & $5.24\times10^{3}$ &
$1.01\times10^{4}$ & $5.34\times10^{3}$ & $\mathbf{7.06\times10}^{2}$\\\hline
Friedman 3 & $2.08\times10^{-2}$ & $1.58\times10^{-2}$ & $1.88\times10^{-2}$ &
$2.87\times10^{-2}$ & $1.93\times10^{-2}$ & $\mathbf{9.76\times10}^{-3}%
$\\\hline
Regression & $1.25\times10^{4}$ & $1.16\times10^{4}$ & $9.93\times10^{3}$ &
$1.24\times10^{4}$ & $9.80\times10^{3}$ & $\mathbf{8.98\times10}^{3}$\\\hline
Sparse & $2.79\times10^{0}$ & $2.20\times10^{0}$ & $1.95\times10^{0}$ &
$2.64\times10^{0}$ & $2.05\times10^{0}$ & $\mathbf{1.44\times10}^{0}$\\\hline
\end{tabular}
\label{t:regression_results}%
\end{table}%

We should point out also that the use of PRGBM reduces the training time in
comparison with the GBM. Table \ref{t:regression_time} shows the training time
for the GBM and the PRGBM, including the time of selecting the tuning
parameters by means of the cross-validation, for four datasets with the
largest number of training examples. The comparison results are presented only
for these two models because they provide the best prediction accuracy. It can
be seen from Table \ref{t:regression_time} that the PRGBM provides the best
time of training for all datasets.%

\begin{table}[tbp] \centering
\caption{Training times of the GBM and the PRGBM}%
\begin{tabular}
[c]{ccc}\hline
Data set & GBM & PRGBM\\\hline
California & $2.2\times10^{2}$ & $\mathbf{4.2\times10}^{1}$\\\hline
HouseART & $3.8\times10^{1}$ & $\mathbf{2.1\times10}^{1}$\\\hline
Boston & $5.1\times10^{0}$ & $\mathbf{3.9\times10}^{0}$\\\hline
Diabetes & $4.1\times10^{0}$ & $\mathbf{3.7\times10}^{0}$\\\hline
\end{tabular}
\label{t:regression_time}%
\end{table}%

\section{Conclusion}

The GBM based on partially randomized trees has been considered in the paper.
The proposed model aims to reduce the computational complexity of the GBM and
to take into account cases when regions of training data are not densely
covered by points. The advantages of the partially randomized trees have been
illustrated by means of synthetic and real data. By studying the proposed
trees, we aimed to extend a large amount of the GBM modifications as well as
the ensemble-based models by an additional model which may improve the
regression accuracy and to reduce the model complexity. Moreover, the use of
the proposed model, we solve the problem of the grid density and inject the
additional randomness which may reduce the overfitting problem which arises
due to greedy of the tree building procedure.

In spite of many numerical examples which show superiority of the proposed
model, the theoretical justification of the obtained results is required. It
can be viewed as a direction for further research. The approach has been
studied for solving the regression problem. At the same time, it is
interesting to consider it also for solving the classification problem. This
is another direction for further research. There are other open questions
related to partially randomized trees, for example, their use in random
forests, in deep forests, etc. These questions can be also regarded as
direction for further research.

\section*{Acknowledgement}

The research was funded as a part of the state assignment for basic research (a code of the research is 0784-2020-0026).


\begin{thebibliography}{10}

\bibitem{Biau-Scornet-2016}
G.~Biau and E.~Scornet.
A random forest guided tour.
{\em Test}, 25(2):197--227, 2016.

\bibitem{Breiman-1996}
L.~Breiman.
Bagging predictors.
{\em Machine Learning}, 24(2):123--140, 1996.

\bibitem{Breiman-2001}
L.~Breiman.
Random forests.
{\em Machine learning}, 45(1):5--32, 2001.

\bibitem{Breiman-etal-1984}
L.~Breiman, J.~Friedman, C.J. Stone, and R.A. Olshen.
{\em Classification and regression trees}.
CRC press, 1984.

\bibitem{Chen-Guestrin-2016}
T.~Chen and C.~Guestrin.
Xgboost: A scalable tree boosting system.
In {\em Proceedings of the 22nd {ACM SIGKDD} International Conference
  on Knowledge Discovery and Data Mining}, pages 785--794, New York, NY, 2016.
  ACM.

\bibitem{Cutler-Zhao-01}
A.~Cutler and G.~Zhao.
Pert-perfect random tree ensembles.
{\em Computing Science and Statistics}, 33:490--497, 2001.

\bibitem{Dietterich-00}
T.G. Dietterich.
An experimental comparison of three methods for constructing
  ensembles of decision trees: Bagging, boosting, and randomization.
{\em Machine Learning}, 40(2):139--157, 2000.

\bibitem{Dorogush-etal-2018}
A.V. Dorogush, V.~Ershov, and A.~Gulin.
Catboost: gradient boosting with categorical features support.
arXiv:1810.11363, October 2018.

\bibitem{Feng-etal-20}
J.~Feng, Y.X. Xu, Y.~Jiang, and Z.-H. Zhou.
Soft gradient boosting machine.
arXiv:2006.04059, June 2020.

\bibitem{Feng-Yu-Zhou-2018}
J.~Feng, Y.~Yu, and Z.-H. Zhou.
Multi-layered gradient boosting decision trees.
In {\em Advances in Neural Information Processing Systems}, pages
  3551--3561. Curran Associates, Inc., 2018.

\bibitem{Ferreira-Figueiredo-2012}
A.J. Ferreira and M.A.T. Figueiredo.
Boosting algorithms: A review of methods, theory, and applications.
In C.~Zhang and Y.~Ma, editors, {\em Ensemble Machine Learning:
  Methods and Applications}, pages 35--85. Springer, New York, 2012.

\bibitem{Freund-Shapire-97}
Y.~Freund and R.E. Schapire.
A decision theoretic generalization of on-line learning and an
  application to boosting.
{\em Journal of Computer and System Sciences}, 55(1):119--139, 1997.

\bibitem{Friedman-2001}
J.H. Friedman.
Greedy function approximation: A gradient boosting machine.
{\em Annals of Statistics}, 29:1189--1232, 2001.

\bibitem{Friedman-2002}
J.H. Friedman.
Stochastic gradient boosting.
{\em Computational statistics \& data analysis}, 38(4):367--378,
  2002.

\bibitem{Geurts-etal-06}
P.~Geurts, D.~Ernst, and L.~Wehenkel.
Extremely randomized trees.
{\em Machine learning}, 63:3--42, 2006.

\bibitem{Guolin-etal-17}
K.~Guolin, M.~Qi, F.~Thomas, W.~Taifeng, C.~Wei, M.~Weidong, Y.~Qiwei, and
  L.~Tie-Yan.
Lightgbm: A highly efficient gradient boosting decision tree.
In {\em Proceedings of the 31st International Conference on Neural
  Information Processing Systems (NIPS'17))}, pages 3149--3157, 2017.

\bibitem{Guryanov-19}
A.~Guryanov.
Histogram-based algorithm for building gradient boosting ensemblesof
  piecewise linear decision trees.
In {\em Proceedings of the 8th International Conference on Analysis
  of Images, Social Networks and Texts. AIST 2019}, volume 11832 of {\em LNCS},
  pages 39--50, Cham, 2019. Springer.

\bibitem{Hastie-Tibshirani-1990}
T.~Hastie and R.~Tibshirani.
{\em Generalized additive models}, volume~43.
CRC press, 1990.

\bibitem{Hastie-Tibshirani-Friedman09}
T.~Hastie, R.~Tibshirani, and J.~Friedman.
{\em The Elements of Statistical Learning}.
Springer, New York, 2009.

\bibitem{Ho-1998}
T.K. Ho.
The random subspace method for constructing decision forests.
{\em IEEE Transactions on Pattern Analysis and Machine Intelligence},
  20(8):832--844, 1998.

\bibitem{Kuncheva-2004}
L.I. Kuncheva.
{\em Combining Pattern Classifiers: Methods and Algorithms}.
Wiley-Interscience, New Jersey, 2004.

\bibitem{Lu-etal-2019}
H.~Lu, S.P. Karimireddy, N.~Ponomareva, and V.~Mirrokni.
Accelerating gradient boosting machine.
arXiv:1903.08708v2, September 2019.

\bibitem{Lu-Mazumder-18}
H.~Lu and R.~Mazumder.
Randomized gradient boosting machine.
arXiv:1810.10158v2, October 2018.

\bibitem{Natekin-Knoll-13}
A.~Natekin and A.~Knoll.
Gradient boosting machines, a tutorial.
{\em Frontiers in neurorobotics}, 7(Article 21):1--21, 2013.

\bibitem{Polikar-2012}
R.~Polikar.
Ensemble learning.
In C.~Zhang and Y.~Ma, editors, {\em Ensemble Machine Learning:
  Methods and Applications}, pages 1--34. Springer, New York, 2012.

\bibitem{Rokach-2019}
L.~Rokach.
{\em Ensemble Learning: Pattern Classification Using Ensemble
  Methods}, volume~85.
World Scientific, 2019.

\bibitem{Sagi-Rokach-2018}
O.~Sagi and L.~Rokach.
Ensemble learning: A survey.
{\em WIREs Data Mining and Knowledge Discovery}, 8(e1249):1--18,
  2018.

\bibitem{Smola-Scholkopf-2004}
A.J. Smola and B.~Scholkopf.
A tutorial on support vector regression.
{\em Statistics and Computing}, 14:199--222, 2004.

\bibitem{Wehenkel-etal-06}
L.~Wehenkel, D.~Ernst, and P.~Geurts.
Ensembles of extremely randomized trees and some generic
  applications.
In {\em Proceedings of Robust Methods for Power System State
  Estimation and Load Forecasting}, pages 1--10, 2006.

\bibitem{Wozniak-etal-2014}
M.~Wozniak, M.~Grana, and E.~Corchado.
A survey of multiple classifier systems as hybrid systems.
{\em Information Fusion}, pages 3--17, 2014.

\bibitem{ZH-Zhou-2012}
Z.-H. Zhou.
{\em Ensemble Methods: Foundations and Algorithms}.
CRC Press, Boca Raton, 2012.

\bibitem{Zhou-Feng-2017a}
Z.-H. Zhou and J.~Feng.
Deep forest: Towards an alternative to deep neural networks.
In {\em Proceedings of the 26th International Joint Conference on
  Artificial Intelligence (IJCAI'17)}, pages 3553--3559, Melbourne, Australia,
  2017. AAAI Press.

\end{thebibliography}

\end{document}